# Development and Analysis of Digging and Soil Removing Mechanisms for Mole-Bot: Bio-Inspired Mole-Like Drilling Robot

Junseok Lee, Christian Tirtawardhana, and Hyun Myung, *Senior Member, IEEE*

*Abstract*— Interests in exploration of new energy resources are increasing due to the exhaustion of existing resources. To explore new energy sources, various studies have been conducted to improve the drilling performance of drilling equipment for deep and strong ground. However, with better performance, the modern drilling equipment is bulky and, furthermore, has become inconvenient in both installation and operation, for it takes complex procedures for complex terrains. Moreover, environmental issues are also a concern because of the excessive use of mud and slurry to remove excavated soil. To overcome these limitations, a mechanism that combines an expandable drill bit and link structure to simulate the function of the teeth and forelimbs of a mole is proposed. In this paper, the proposed expandable drill bit simplifies the complexity and high number of degrees of freedom of the animal head. In addition, a debris removal mechanism mimicking a shoulder structure and forefoot movement is proposed. For efficient debris removal, the proposed mechanism enables the simultaneous rotation and expanding/folding motions of the drill bit by using a single actuator. The performance of the proposed system is evaluated by dynamic simulations and experiments.

## I. INTRODUCTION

As the depletion of existing energy sources such as oil and coal approaches, new research and technological advancements are essential to explore new energy sources. Resource exploration contributes to the advancement in industries, and for the procurement of newer resources, technologies such as new drilling systems are, therefore, crucial. Accordingly, several sophisticated rotary drill bits and stable directional drilling mechanisms have been proposed [1-2]. On the other hand, the size and weight of the systems have increased to achieve the goal. Shale gas and coalbed methane, relatively new energy sources, are located in rigid geological formations usually distributed horizontally within the same layer. To explore these resources, it is desirable to apply directional drilling techniques to reduce drilling time and costs [3]. In addition, scientists worldwide are in search of rare earth element (REE) deposit areas since they are continuously increasing in demand due to the development of electronic instruments. As a result of several explorations, it has been confirmed that REEs can be found in shallow locations approximately within 30 m depth below the ground. Moreover, the layer in which REEs are distributed is a soil group with high iron oxide content, mainly called the laterite layer, which has a low compressive strength with a maximum strength of approximately 1.10 MPa [4-5]. There is a need for drilling systems suitable for shallow depths and soft grounds; e.g. efficient REE exploration systems. In addition, in recent years, the development of drilling systems for planetary exploration has been in demand, and several kinds of studies have attempted to study the lunar soil in extreme environments [6]. The planetary drill system, developed by Honeybee Robotics, uses drilling bits and prismatic movements and can excavate up to 50 feet [7]. However, heavy and bulky frames are used to maintain stability. In another study, the Planetary Underground Tool (PLUTO) was proposed for collecting Martian soil [8]. Unlike the planetary drilling system, PLUTO uses hammers for drilling and can drill up to 1.5 m with a speed of 0.75 meters per hour. Other planetary drill system for Venus exploration can reach a speed of 0.35 meters per hour [9-10]. For efficient removing and transporting soil, Self-Turning Screw Mechanism (STSM) is developed [11-12]. However, it cannot drill deeply due to the limitation of its rotary power. Consequently, existing systems are not effective for excavation due to various limitations, such as low Rates of Penetration (ROP) and limited space to support heavy and large equipment. In addition to the limitations of soil and rock drilling technologies, the treatment of excavated soil causes several side effects, including environmental pollution because of the use of fluids such as mud or slurry. Debris is transported through the flow of fluids to the inlet and outlet of the drill hole, but an additional filtration system is required. This approach is not suitable for use in weak grounds since the weak grounds cannot endure high pressure from the fluid when transporting the soil. Recently, various studies and developments on bioinspired robots that simulate the body structure and habits of animals have been conducted. In this paper, to overcome the limitations of previous systems, excavation and debris removal mechanisms that mimic animals with excavation habits are proposed. Animals with excavation habits can efficiently excavate and move large amounts of soil relative to their small size. Their unique skills are derived by their inherent biological structure and habits. Therefore, an excavation robot that mimics the structure and behavior of these animals is also expected to be efficient.

*This work is supported by the Technology Innovation Program (#10076532, Development of embedded directional drilling robot for drilling and exploration) funded by the Ministry of Trade, Industry & Energy (MOTIE, Korea). The students are supported by Korea Ministry of Land, Infrastructure and Transport (MOLIT) as 「Innovative Talent Education Program for Smart City」.

Junseok Lee is a Ph.D. candidate at the School of Electrical Engineering, Korea Advanced Institute of Science and Technology (KAIST), Korea (e-mail: ljs630@kaist.ac.kr).
Christian Tirtawardhana is a researcher at the School of Electrical Engineering, Korea Advanced Institute of Science and Technology (KAIST), Korea (e-mail: christian@kaist.ac.kr).
Hyun Myung is a Professor at the School of Electrical Engineering and the Head of Robotics Program, also with KI-AI and KI-R, Korea Advanced Institute of Science and Technology (KAIST), Korea (e-mail: hmyung@kaist.ac.kr).

## II. ANATOMY AND DIGGING HABIT OF A MOLE

Animals with excavation habits have specialized excavating abilities to build and live in nests under the ground. These abilities are associated with different excavating methods that have evolved depending on the habitat environment [13]. Mole-rats mainly use their teeth to scrape soil into clods and then divide the soil more loosely, making the removal of the soil easier. The moles have a powerful grip and can crush the soil using two incisors and a large jaw [14]. The moles have an average weight of approximately 40 g and a maximum weight of only 85 g [15]. However, they can exert approximately 48 times larger grinding forces than the average body weight. In addition to having strong teeth and jaws, the moles can freely move their heads, which provides many advantages in excavation. The distance between the upper and lower incisor is only approximately 20 mm at maximum [16]. However, they can excavate holes with a width of approximately 50 ± 1 mm [17-18]. Despite the limited range of their teeth, the freedom of movement in the head makes it possible to form holes much larger than their body size.

Another species, the European mole, uses two large forelimbs to simultaneously excavate and remove debris. These moles have a special bone structure that enables the large digging force of the forefoot. The major bones involved in forefoot movement consist of the scapula, humerus, and clavicle [19]. Humans have wide, flat, and rounded scapula for a wide range of shoulder motions. On the other hand, the mole has a different structure. The mole's shoulder joint can only be rotated by 10-15°, allowing them to push the soil sideways. A mole's scapula has a very thin and long special structure and a broad area. As a result, several muscles are attached and allow them to dig with large forces [20]. When the muscles pull the scapula, a rotational movement occurs in the humerus, and the entire arm rotates to allow movement in the forelimbs. The humerus, on the other hand, is connected with the rotational joint at the shoulder and stabilizes the movement of the arm. A human humerus has a long, thick tubular shape, thus allowing a wide range of motions of the arm. In contrast, the humerus of the mole is flat and broad, like a plate. These short and dense humerus bones are able to transmit higher torques to the forelimbs when the moles are digging and are well suited for moving in hard and tight spaces inside a borehole. Another important bone is the clavicle, which supports the shoulders reliably [21]. The clavicle supports high torques of the forelimbs and transmits the force efficiently despite its small size. This paper aims to develop a system that can excavate shallow and soft grounds. Therefore, an efficient excavation and debris removal mechanism using a hybrid excavation mechanism that combines properties of a mole-rat's teeth and jaw and a European mole's forelimb structure is proposed.

## III. BIO-INSPIRED DESIGN OF DIGGING AND SOIL REMOVING MECHANISMS

### A. Expandable Drill Bit

Existing expandable methods follow similar mechanisms to those proposed by [22-23] to operate the inner shaft and achieve scalability. However, the problem is that these methods require bulky high-pressure hydraulic pistons for their operation. As a result, additional large equipment is required to drive the developed expandable mechanisms. This approach is not desirable in shallow excavation applications and small robots. Therefore, this paper proposes a new expandable mechanism that can be driven using only one small actuator. To that end, as mentioned above, a bioinspired expandable drill bit that can simultaneously realize the mole-rat incisor structure and the high degrees of freedom of the mole-rat head is proposed. The expandable drill bit consists of three main parts: the inner shaft, middle part, and outer case. The rotating motor is connected only to the inner shaft, and the outer case is fixed to the robot body and does not rotate. The inner shaft and the middle part are connected in a linear pattern so that when the inner shaft rotates, the middle part also rotates at the same speed. The middle part and the outer case are connected by a screw pattern. Because the outer case is fixed, when the inner shaft is rotated by the rotation of the motor, the middle part moves linearly up and down through the screw pattern. The middle part moves forward when the motor rotates clockwise and moves backward when the motor rotates counterclockwise. The schematic of this system is shown in Fig. 1.

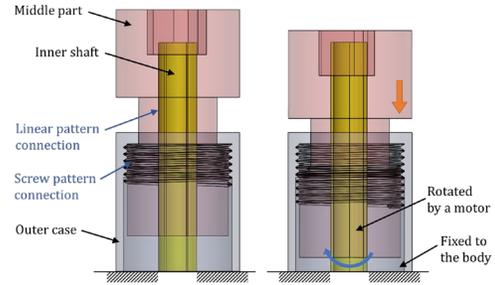

Fig. 1. Screw pattern and linear motion of the expandable drill bit.

The expansion and contraction mechanism of the blade is shown in Fig. 2. Blades that come in direct contact with the ground are classified as inner and expandable blades. The expandable blade functions to expand and contract according to the linear movement of the body by the rotation of the motor. There are three expandable blades and three inner blades, where each blade is arranged in uniform intervals. Rack and pinion systems are applied to implement the expansion and contraction mechanisms of the expandable blades, which are key parts for drilling. Three racks are vertically coupled to the inner shaft. Each expandable blade has a gear-shaped segment, and each segment is connected with a rack. Therefore, when the middle part is reversed, the gear rotates outward and the blade expands. In the opposite case, the middle part moves forward and the gear rotates inward and contracts. In our case, the system is capable of drilling 93.4 mm in diameter in the folded state and 202 mm in diameter in the expanded state, with an expandability of 2.16 times to the folded state.

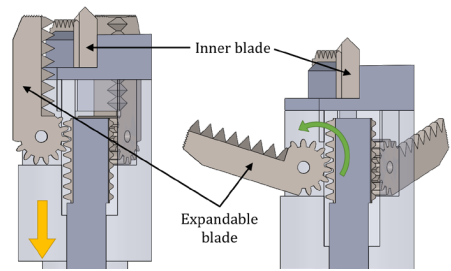

Fig. 2. Expansion mechanism of the expandable blades.

Consequently, it is expected that during excavation, the proposed expandable drill bit will be advantageous for controlling the robot's posture in a borehole by mimicking the mole-rat incisors and the wide range of motion of the mole-rat's jaw.

### B. Caster Wheel Mechanism

An expandable drill bit is designed to excavate holes with a larger diameter than the body of the robot and to excavate when the blades are fully extended. When the blades are fully extended, the rotation of the motors that control the expansion should be restricted to ensure that the blades do not go over 90 degrees. Since the motor mentioned above has a limitation to the number of rotations it can make for expansion of the blades, it cannot function simultaneously as a drilling motor. Therefore, a unique mechanism that allows continuous rotation while limiting the linear displacement of the middle part is applied. The proposed mechanism is based on two points. The first is to limit the length of the thread that exists between the middle part and the outer case. Therefore, the linear motion of the middle part depends on the length of the thread connected with the outer case. If the screw pattern of the middle part is completely moved out from the outer case by successive rotations, the linear motion will be interrupted. However, this method can cause durability problems by causing the middle part's thread to continuously impact the end of the outer case's thread during continuous rotations. Therefore, the second key point is to avoid thread-to-thread conflicts when drilling in the expanded state. Thus, a set of six caster wheels are inserted in a circular pattern in the middle part. The designed caster wheel sets and drill bit are shown in Fig. 3. The wheels touch the inner surface of the outer case and revolve during rotations. Each wheel set is mounted at an orientation of $\theta$ (e.g. 30°) and can be rotated to a horizontal state. The spring is mounted to the caster in the opposite direction of the drill rotation. The difference between the heading and travel directions generates a cornering force to the wheels, resulting in a self-aligning torque [24]. If this self-aligning torque is greater than the spring's elasticity, the spring is compressed and the wheel rotates in a direction perpendicular to the drilling direction. By applying this mechanism, continuous rotation can be realized without excessive friction or impact on the thread of the middle part and the outer case. In the case of counter-rotation, the spring is retracted, and the wheel is returned to its original position because the self-aligning torque is limited. As the system continues to rotate clockwise, the friction of the wheels creates a forward force on the middle part. This additional force allows the middle part to securely engage the thread of the outer part and move forward with blade reduction. This mechanism uses two wheels for each caster set to produce sufficient friction and torque.

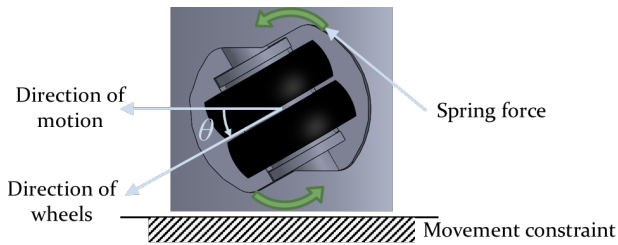

Fig. 3. Wheel structure.

### C. Soil Removing Mechanism

For removing excavated soils, a bioinspired forelimb mechanism for debris removal based on the scapula and humerus interaction of a mole is proposed. The basic mechanism of these natural diggers is the exertion of the scapula to allow the conversion of linear pulling motions generated by the attached muscles to robust rotational forces at the humerus. The long scapula structure is connected to a linear actuator that is fixed to the body of the robot. As shown in Fig. 4, the linear motor pulls the scapula backward, causing the humerus to rotate, resulting in the rotation of the forefoot. Moreover, the proposed design uses a pair of wide and flat forefeet, and the end of the tip is curved to fit the intended cross-sectional shape of the drilled hole.

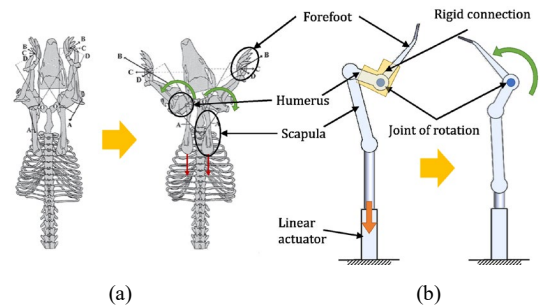

Fig. 4. Comparison of forelimbs. (a) Forelimbs structure of the natural diggers. (b) Bioinspired forelimb mechanism.

However, the described mechanism implements only the rotational movement of the forefoot. Compared with a mole's actual behavior, the proposed mechanism only allows rotation of the forefoot but is limited in its flexibility for contracting and pushing back debris. Therefore, there is an additional need for a mechanism that can realize linear motion. The further design combines a worm gear and rack-pinion mechanisms to achieve this movement. The complete model which combines the expandable drill bit and forelimb structure is shown in Fig. 5.

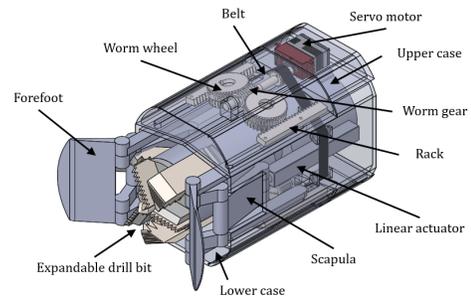

Fig. 5. Whole structure of the digging system (forelimb structure with expandable drill bit).

One servo motor used as a power source is installed in the center of the upper case and connected to the worm gear. The worm gear controls the rotation of two worm wheels connected to both sides. By the configuration above, the rotational motion of the motor distributes power symmetrically. Each link is coupled to a rack that translates as the worm wheel moves. This mechanism converts the rotational movement of the worm gear into the linear movement of the forelimbs. The same configuration, without the servo motor, is placed at the bottom of the case to balance the movements, and the belt distributes power to the bottom to synchronize the movements. The system controls the rotation of the servo motor clockwise and counterclockwise to move the forelimb joint back and forth, respectively. In this system, the difference between the linear velocity of the scapula and the linear velocity of the rack connected to the forelimbs determines the movement. If both cases operate at the same speed, the forelimb moves linearly forward or backward. When the forelimb moves forward, the forelimb generates a spreading motion when the linear velocity of the scapula is less than the rotational joint speed of the forelimb. In contrast, when the forelimb moves backward, the forelimb generates a gathering motion.

### D. Digging Sequence of the System

For the excavation of the robot, a digging sequence combining the proposed mechanism is required. The modules are operated alternately to avoid collisions of the drill bit and the forelimbs. The digging sequence of the system is shown in Fig. 6.

(1) Excavation begins with the forelimbs located at the rear of the robot, and the folded drill bit advances forward. (2) Then, the drill bit rotates counterclockwise, expanding the blades. (3) The extended drill bit continues to rotate, contacting the surface and crushing the soil. (4) After a sufficient amount of drilling is completed, the drill bit rotates clockwise to retract the blade and returns to its initial position. (5) The blade retracts to provide a sufficient amount of space for the forelimbs to move forward after gathering both end effectors. (6) Finally, the scapula is pulled to spread the forefoot to the left and right, and the forelimbs move backward to remove debris. The excavated soil is removed on either side of the robot, and this process prevents the robot from interfering with forward movement. The removed soil accumulates on the rear path that the robot has passed.

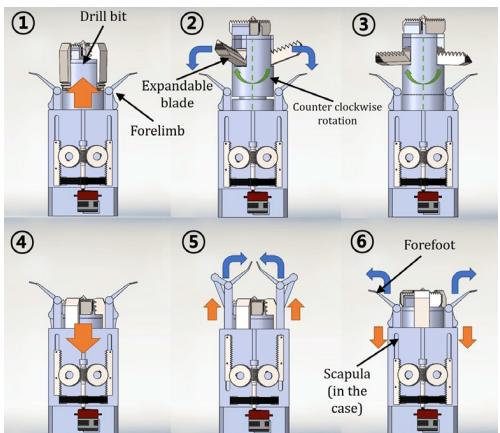

Fig. 6. Digging sequence diagram of the system.

## IV. SYSTEM ANALYSIS

### A. Analysis of Drill Bit by Simulation and Experiment

In the proposed design, two metallic elements are used, i.e., steel ($\sigma_{y,s}$ = 460 MPa) for the blades, rack-pinion sets, and worm gear sets; and aluminum ($\sigma_{y,a}$ = 75 MPa) for the remaining parts. The targeted soil property is set to have a maximum compressive strength, $\sigma_c$, of 4 MPa since the common mixture of soils in shallow locations has a compressive strength less than 2.5 MPa [25]. Since soils have a fairly high degree of uncertainty, safety factors should be applied for conservative design. Therefore, this study used a compressive strength of 4 MPa as the excavation target. To simplify the complexity of the soil model, the properties of Autoclaved Lightweight Concrete (ALC) are employed. Table I shows the specific properties of materials for simulation, and Table II shows the coefficients of friction between the materials used [26].

TABLE I. MATERIAL PROPERTIES FOR THE SIMULATION

| Material | Properties | Values | Unit |
|---|---|---|---|
| Steel | Young's modulus, $E_s$ | 200,000 | MPa |
| | Tensile strength, $\sigma_{t,s}$ | 560 | MPa |
| | Yield strength, $\sigma_{y,s}$ | 460 | MPa |
| | Poisson's ratio, $\nu_s$ | 0.29 | - |
| | Mass density, $\gamma_s$ | 7,800 | kg/m³ |
| Aluminum | Young's modulus, $E_a$ | 69,000 | MPa |
| | Tensile strength, $\sigma_{t,a}$ | 85 | MPa |
| | Yield strength, $\sigma_{y,a}$ | 75 | MPa |
| | Poisson's ratio, $\nu_a$ | 0.33 | - |
| | Mass density, $\gamma_a$ | 2,700 | kg/m³ |
| Concrete | Young's modulus, $E_c$ | 1,715 | MPa |
| | Compressive strength, $\sigma_c$ | 4 | MPa |
| | Poisson's ratio, $\nu_c$ | 0.2 | - |
| | Mass density, $\gamma_c$ | 650 | kg/m³ |

TABLE II. COEFFICIENT OF FRICTION BETWEEN MATERIALS

| Material 1 | Material 2 | Coefficient of friction (dry) | |
|---|---|---|---|
| | | Static, $\mu_s$ | Kinetic, $\mu_k$ |
| Steel | Steel | 0.78 | 0.42 |
| Steel | Aluminum | 0.61 | 0.47 |
| Steel | Concrete | 0.57-0.75 | 0.45 |
| Aluminum | Aluminum | 1.05-1.35 | 1.40 |

Two simulations are performed to evaluate the performance and feasibility of the proposed drilling system.

First, the numerical model is used to evaluate the interaction between the drill bit and the soil. This simulation is performed using finite element analysis software for dynamic explicit simulations. Two components are constructed: a drill bit and a cylindrical testbed representing the soil to be excavated. The numerical analysis is divided into two parts: inner drilling related to inner blades and external drilling for evaluating the performance of the expandable blades. In this simulation, the material properties described in Tables I and II are used.

The drill bit only drills up to a depth of 30 mm in one excavation session in the system, and it is estimated that a drilling duration of 86 seconds at approximately 120 rpm is required. The results of the drilling simulation for the inner blades and expandable blades are shown in Fig. 7. The simulations confirm that the soil can be successfully excavated by generating stresses above the compressive strength of the specimen, 4 MPa. The graphs in Fig. 8 shows the required changes in torque over time and the changes in weight on bit (WOB) over time according to bit type. Based on the simulation, for 30 mm deep drilling, the maximum required torque ($\tau_{R-max}$) is 5.17 Nm, and the required weight on the bit ($WOB_{R-max}$) is 106.95 N.

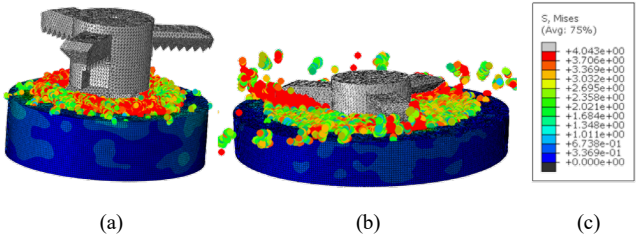

(a)          (b)          (c)

Fig. 7. Numerical simulation results between the drill bit and ALC block. (a) Inner blade drilling. (b) Expandable blade drilling. (c) Color spectrum corresponding to stress intensity.

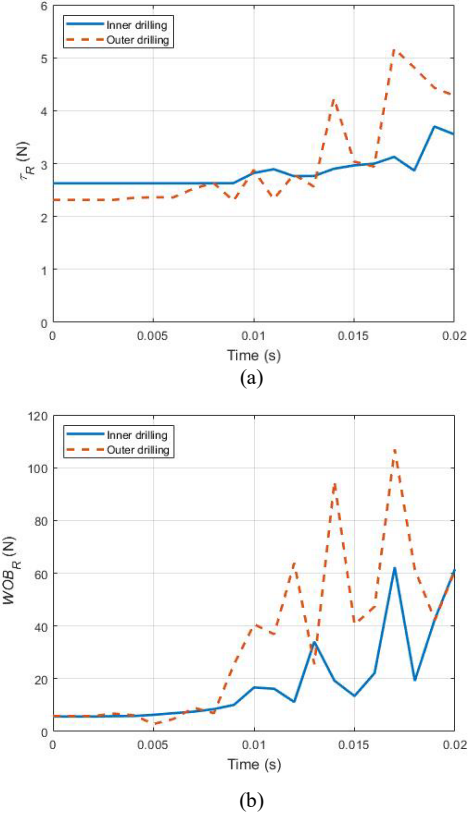

Fig. 8. (a) Changes in torque required over time for the inner drill bit and expandable drill bit. (b) Changes in weight on bit required over time for the inner drill bit and expandable drill bit.

Second, the capacity of the drill bit structure based on the force and torque values generated by the drill bit system is evaluated using finite element analysis and a static simulation. To evaluate the response of the drill bit structure in the worst loading scenario, it is assumed that all input forces are absorbed completely by the body rather than being transferred to linear motion. Thus, in this simulation, the expandable blades are fixed in the extended state. The maximum values of $\tau_{R-max}$ and $WOB_{R-max}$ are applied to the surface of the inner shaft, which is connected to the motor. The maximum stress $\sigma_{max}$ occurs as 4.682 MPa in the inner shaft, which is much lower than the stress capacities of steel and aluminum, which are 560 MPa ($\sigma_{t,s}$) and 85 MPa ($\sigma_{t,a}$), respectively. Therefore, through this series of evaluations, the drilling performance and the structure of the drill bit are verified.

To verify the drilling capability of the system, two analyses are performed, i.e., the drilling performance of the blades and the wheel mechanism for continuous drilling. Teale [27-28] established the specific energy, $E_s$ for rotary drilling, which can be derived as follows:

$$E_s = \frac{WOB}{A_B} + \frac{2\pi \cdot RPM \cdot T}{A_B \cdot ROP} \quad (1)$$

where $RPM$ is the rotation speed of the drill bit, $A_B$ represents the contact area of the bit and $T$ is the torque of the bit. The torque can be calculated as follows:

$$T = \mu \cdot \frac{D_B \cdot WOB}{3} \quad (2)$$

where $\mu$ ($\approx$ 0.45) is the friction coefficient between the drill bit and the targeted soil and $D_B$ ($\approx$ 202 mm) is the expanded diameter of the drill bit. Galle demonstrated the effect of WOB, RPM, and the dullness of the bit to the rate of penetration (ROP) through the following semi-empirical equation [29]:

$$ROP = C_f \cdot \frac{\overline{WOB}^k \cdot r}{a^P} \quad (3)$$

where $C_f$ represents the formation drilling capability parameter (can be estimated as $\overline{WOB}^{0.6}$), $a$ is a function of dullness ($\approx$ 565.6), $k$ is the exponent on weight (in most formations, equal to 1), and $P$ is the exponent on the function of dullness (equal to 0.5) for the chipping type bit tooth wear. Additionally, $r$ denotes the function of essential rotational speed, expressed as a fractional exponent, as follows [29]:

$$r = \begin{cases} e^{\frac{-100}{RPM^2}} \cdot RPM^{0.428} + 0.2RPM(1 - e^{\frac{-100}{RPM^2}}) \\ \qquad\qquad\qquad\qquad\qquad\text{(soft condition)} \\ e^{\frac{-100}{RPM^2}} \cdot RPM^{0.750} + 0.5RPM(1 - e^{\frac{-100}{RPM^2}}) \\ \qquad\qquad\qquad\qquad\qquad\text{(hard condition)} \end{cases} \quad (4)$$

$$\overline{WOB} = \frac{7.88 \cdot WOB}{D_B} \quad (5)$$

In this paper, the target strength of the soil is about 4 MPa. Therefore, formation of $r$ is considered as soft condition.

The motor used in this research has a stall torque, $\tau_S$, equal to 8.83 Nm and no-load speed, $\omega_n$, of 200 rpm. Thus, the relationship between the torque and RPM of the motor can be defined using the following equation [1]:

$$RPM = \frac{(\tau_S - \frac{T}{\eta}) \cdot \omega_n}{\tau_S} \quad (6)$$

where $\eta$ ($\approx$ 0.84) is the transfer efficiency between the motor and drill bit.

The excavation experiment is conducted using the manufactured drill bit. The experiment for drilling is conducted in the same condition as the simulation. The actual drill bit module is made of the steel and aluminum as the simulation and drills the ALC block. An acrylic pipe is used as the guide to fix the drill bit, and the average compressive strength of the block is measured to be approximately 4 MPa. The motor used for the drilling module is a DC motor geared with a reduction ratio of 15:1. The input voltage supplied from the power supply is 24 V and the maximum speed is about 200 rpm. The measurement for drilling depth and RPM is performed for 10 minutes after the expandable blade touches the ALC block. The test is carried out under various WOB conditions, and the fundamental weight ($W$) of the drill bit and the motor is 7 kg. The test environment is shown in Fig. 9, and the results are shown in Table III.

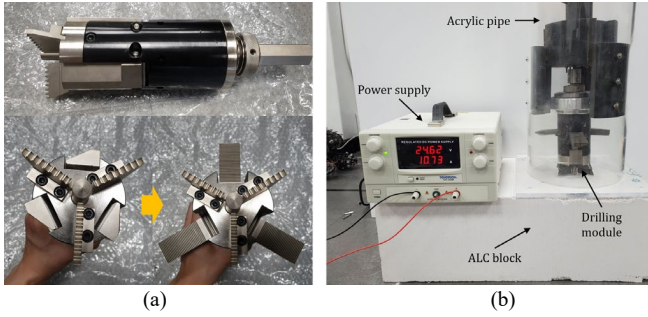

Fig. 9. (a) Structure of expandable drill bit. (b) Experiment environment for drilling ALC block.

TABLE III. RESULTS OF DRILLING EXPERIMENT ACCORDING TO WOB. $W$ DENOTES THE FUNDAMENTAL WEIGHT.

| WOB (kg) | $W$ | $W$ + 0.5 | $W$ + 1.0 | $W$ + 2.0 | $W$ + 5.0 |
|---|---|---|---|---|---|
| Drilled depth (mm) | 91.09 | 16.63 | 122.57 | 149.78 | 233.31 |
| RPM (rev/min) | 124 | 110 | 190 | 81 | 74 |
| $E_s$ (MPa) | 6.58 | 6.12 | 5.60 | 5.07 | 4.17 |

As shown in Table III, it is confirmed that RPM and $E_s$ tend to decrease as WOB increases. In contrast, the drilled depth increases, which means that the ROP is increased. Based on the results in Table III, equations (1) – (6), and the relationship between ROP and WOB obtained by the numerical model, the prediction of the change in specific energy and the ROP of the proposed system by the WOB is shown in Fig. 10. Simulation results (solid lines) are compared with actual experimental results (x marks) to verify drilling performance. The dotted line in Fig. 10 (a) means the compressive strength of the ALC block. The actual experimental result shows a similar tendency to the simulation, and it is estimated that the excavation of 4 MPa soil will have a ROP performance of about 1.05 (diamond mark). In Fig. 10 (b), as the minimum specific energy, $E_{S-min}$, is considered as the compressive strength of the targeted soil, the optimal ROP of the system is estimated to reach 1.05 m/hr when the WOB is 93.3 N and the RPM is 124.49. In addition, the ROP, according to the WOB of the existing drilling system, is marked and compared with the proposed system. The evaluation results show that the proposed system has the highest performance compared to the conventional systems including PLUTO, STSM, and VENUS.

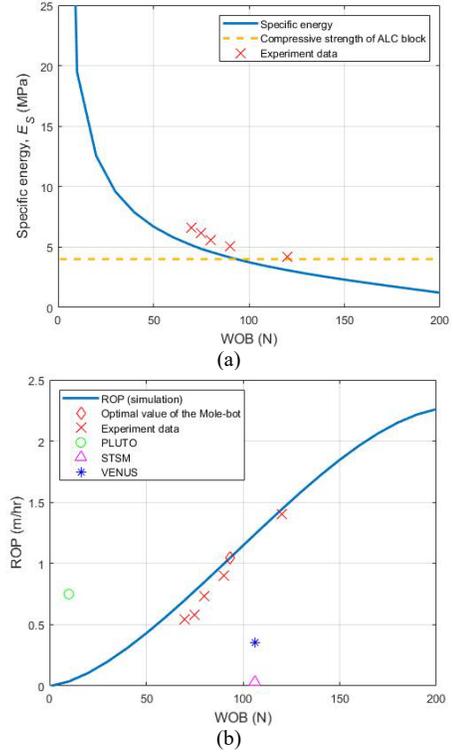

Fig. 10. Change in (a) specific energy and (b) ROP by WOB. The x marks denote experiment results and the diamond mark denotes the expected optimal condition of the proposed drilling system.

### B. Analysis of Wheel Mechanism

To validate the performance of the wheel system in terms of its ability to allow smooth continuous rotation for drilling, a computational analysis is performed. The force equilibrium on each castor wheel is depicted in Fig. 11.

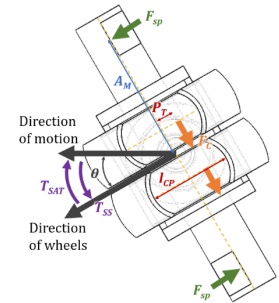

Fig. 11. Working forces on the castor wheel system.

The inclined angle of the castor wheels, $\theta$, is designed to be equal to 30°. According to Hooke's Law, the retaining force of a spring, $F_{sp}$, that maintains the orientation of the castor can be calculated as follows [24]:

$$F_{sp} = k \cdot \Delta x \qquad (7)$$

where $k$ is the spring constant (= 0.077 N/mm) and $\Delta x$ (= 5 mm) is the distance from 30° to 0° angle of the castor orientation. The torque equilibrium of the castor, $\sum T$, at the pivot point can be derived through the following equations:

$$T_{SAT} = 2(F_C \cdot P_T) \qquad (8)$$

$$T_{SS} = 2(F_{sp} \cdot A_M) \tag{9}$$

$$\Sigma T = T_{SS} - T_{SAT} \tag{10}$$

where the cornering force, $F_C$, (= 2.4 MPa) is obtained through a dynamic model of the castor system when $\mu_s$ and $\mu_k$ are equal to 0.9 and 0.75, respectively. Pacejka [31] suggested that the pneumatic trail, $P_T$, can be estimated as a quarter of the contact patch length, $l_{CP}$ (= 7 mm), and the moment arm of the spring force, $A_M$, is equal to 10 mm. In this study, the self-aligning torque, $T_{SAT}$, and torque to correct the slip angle steer, $T_{SS}$, are computed as 0.0084 Nm and 0.0077 Nm, respectively, and represented in Fig. 11. The negative sign of $\Sigma T$ proves that the wheels are able to rotate to 0° orientation, enabling smooth rotation of the system during continuous drilling. Fig. 12 shows the simulation results of the caster wheel system.

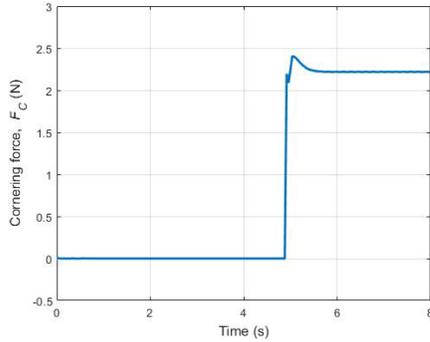
Fig. 12. Simulation results of the caster wheel system. Change in cornering force over time when the castor system moves left in Fig. 11. After contact with the restraining wall, the cornering force increases to a maximum.

### C. Analysis of the Forelimbs Mechanism

The forelimb apparatus is designed to transport soil waste every 30 mm as the mole-bot drills, so the weight of the waste, $w_{sw}$, to be removed amounts to 7.55 N. In this analysis, the evaluation will be in two parts: the evaluation of the pushing mechanism and that of the pulling mechanism, which are performed using a dynamic model of the forelimbs. To evaluate the pushing mechanism, a static hollow box with a wide range of inner width, $d$, (40 mm, 80 mm, 120 mm, 160 mm, and 200 mm) is used as the testbed. Correspondingly, the results of the evaluation show various forefoot opening angles, $\alpha$, measured from the horizontal line. The pushing motion is performed purely by the movement of the linear actuators, where each has a maximum force, $F_m$, of 80 N. From the dynamic simulation, the maximum pushing force in each condition is presented in Table IV, and the illustration of the simulation of the forelimb mechanism is shown in Fig. 13.

TABLE IV. MAXIMUM HORIZONTAL CONTACT FORCE ACHIEVED IN VARIOUS TESTBED WIDTHS FOR THE PUSHING EVALUATION

| $d$ (mm) | $\alpha$ (°) | $F_{H-max}$ (N) |
|---|---|---|
| 40 | 57 | 35.56 |
| 80 | 76 | 38.26 |
| 120 | 94 | 39.90 |
| 160 | 112 | 39.98 |
| 200 | 135 | 33.93 |

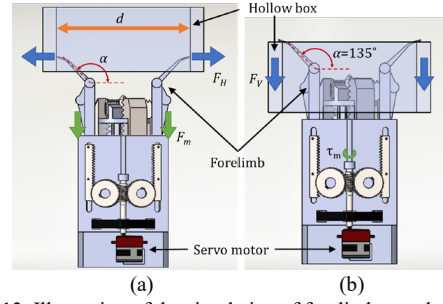
Fig. 13. Illustration of the simulation of forelimbs mechanisms. (a) Pushing evaluation. (b) Pulling evaluation.

Both servomotor (maximum torque, $\tau_m$, of 2.5 Nm) and linear actuators contribute to the pulling motion; however, to simplify the calculation, separate evaluations of the servomotor and linear actuators are performed. The forefeet are locked in a fully open condition ($\alpha = 135°$). The contact forces are presented in Fig. 14 (a), and the resulting pulling forces obtained through a dynamic simulation of both the servomotor and the linear actuator are shown in Fig. 14 (b).

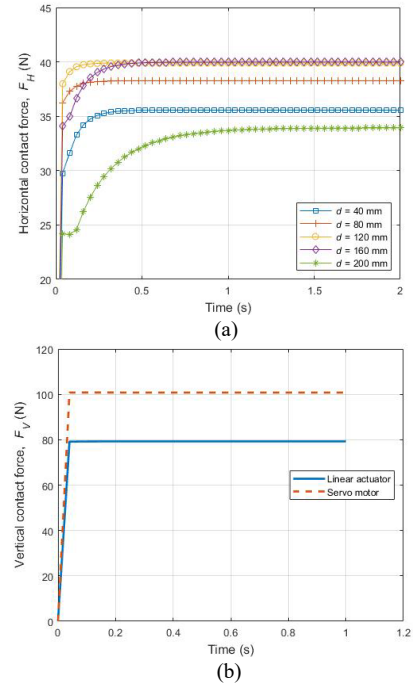
Fig. 14. Simulation analysis results of the forelimbs. (a) Change in the horizontal contact force over time when the forefeet are rotated by the pulling motion of the linear actuator. (b) Change in the vertical contact force over time when the forelimbs are pulled backward by the motion of the servomotor and the linear actuators independently.

The force generated by the servo motor is observed to be much higher than that generated by the linear actuators. The pushing performance depends entirely on the movement of the servo motor, and the linear actuator improves the system efficiency by ensuring a purely vertical pushing action only. Based on these analyses, it can be concluded that the proposed system is able to accomplish the goal of transporting soil waste.

## V. Conclusion

In this study, inspired from burrowing animals, a novel excavation system appropriate for shallow depths and soft soils is proposed. The bioinspired design implements an expandable drill bit and forelimb mechanisms, combining the advantages of the excavation methods of both chisel tooth diggers and humeral rotation diggers. In the proposed design, the expandable drill bit can expand up to 2.16 times the folded size. A novel wheel mechanism built into the drilling system is also proposed, which allows continuous drilling to be performed simultaneously with the expanding/folding of the blades using only one actuator. In addition, the forelimb mechanism, which simulates the anatomy of humeral rotation diggers, enables efficient removal of excavated soil without the need for additional equipment. A series of excavation sequences is designed to mimic the excavation behavior of burrowing animals. The validity of drilling performance is verified using a dynamic explicit simulation with a numerical model of the drill bit and soil. The proposed wheel mechanism and forelimb mechanism are also evaluated using a dynamic simulation. In addition, the actual experiment was conducted and showed a similar tendency with the simulation results. The system is proven to be able to excavate soils with a compressive strength of up to 4 MPa with a borehole diameter of approximately 202 mm. The optimal ROP of the proposed drilling system is about 1.05 m/hr, which overcomes that of the existing drilling systems (less than 1 m/hr), when the WOB is 93.3 N and the RPM is 124.49 rev/min. In addition to the excavation and debris removal modules proposed in this study, locomotion and locking mechanisms are needed for the entire robot to travel and support the excavation reaction forces by itself. After those processes are added to the robot, a real field test with a complete excavation procedure should be carried out. It is expected that the small-scale embedded excavation robot will be applied to the exploration of underground resources and planetary exploration.